\documentclass{article}

\usepackage[final]{neurips_2024}
\usepackage{graphicx}
\usepackage{amsmath}
\usepackage{multirow}
\usepackage{amssymb}
\usepackage{booktabs}
\usepackage{url}

\usepackage{booktabs}
\usepackage[utf8]{inputenc} 
\usepackage[T1]{fontenc}    
\usepackage{hyperref}       
\usepackage{url}            
\usepackage{booktabs}       
\usepackage{amsfonts}       
\usepackage{nicefrac}       
\usepackage{microtype}      
\usepackage{xcolor}         

\title{Assessing Bias in Metric Models for LLM Open-Ended Generation Bias Benchmarks}

\author{
  Nathaniel Demchak$^{1,2}$\textbf{\thanks{\textbf{These authors contributed equally to this work.}}}  ,Xin Guan$^{1}$\textbf{\footnotemark[1]} , Zekun Wu$^{1,3}$\textbf{\footnotemark[1]}  , \textbf{Ziyi Xu}$^{3}$\\\textbf{Adriano Koshiyama}$^1$, \textbf{Emre Kazim}$^1$\\\\
  $^1$ Holistic AI, $^2$Stanford University,  $^3$University College London
}

\begin{document}

\maketitle

\begin{abstract}

Open-generation bias benchmarks evaluate social biases in Large Language Models (LLMs) by analyzing their outputs. However, the classifiers used in analysis often have inherent biases, leading to unfair conclusions. This study examines such biases in open-generation benchmarks like BOLD and SAGED. Using the MGSD dataset, we conduct two experiments. The first uses counterfactuals to measure prediction variations across demographic groups by altering stereotype-related prefixes. The second applies explainability tools (SHAP) to validate that the observed biases stem from these counterfactuals. Results reveal unequal treatment of demographic descriptors, calling for more robust bias metric models.

\end{abstract}

\paragraph{Introduction} Bias in large language models poses significant concerns \citep{BiasDanger}, with numerous benchmarks and metrics proposed for evaluation \citep{biasSurvey, stereoset, crowspairs, chem101}. A common approach involves generating responses and using NLP classifiers as metric models, as demonstrated by BOLD \citep{bold} and SAGED \citep{Saged}. However, these classifiers often carry inherent biases that can skew bias assessments, as highlighted by SAGED. To address this issue, our study systematically examines these biases by evaluating how various bias metrics respond to stereotypes across demographic groups using counterfactuals and popular classifiers. Additionally, we employ explainability methods, such as SHAP, to analyze the contribution of specific demographic descriptor to classifier predictions, providing deeper insights into the underlying causes of observed biases.

\paragraph{Methodology}
We assessed the responses of various bias-detection models to stereotypes by creating counterfactual sentences using the MULTI-GRAIN STEREOTYPE DATASET (MGSD) \citep{Stereotype}. First, we filtered the dataset to include only sentences labeled as “stereotype” and then modified them by prepending demographic descriptors correlated with each stereotype type in MGSD: race (“Whites,” “Blacks,” “Asians”), gender (“Males,” “Females,” “Non-binaries”), profession (“Construction workers,” “Bankers,” “Doctors,” “Nurses,” “Teachers”), and religion (“Atheists,” “Christians,” “Jews,” “Muslims”). This ensured relevance and consistency, as certain stereotypes related to one category (e.g., race) may not apply to another (e.g., profession). Then, Each modified sentence was paired with one of four randomly selected stereotype-related prefixes—“often assume that,” “usually think that,” “sometimes feel that,” and “mostly say that”—using a fixed random seed (42) for reproducibility. For instance, “The Finnish man was very energetic” was transformed into “Males/Females/Non-binaries usually think that the Finnish man was very energetic.” These counterfactuals were then evaluated using four common bias metric models: Detoxify \citep{Detoxify}, Regard \citep{regard}, DistilBERTSentiment \citep{lik_xun_yuan_2023}, and vaderSentiment \citep{Vader}. Next, SHAP (SHapley Additive exPlanations) \citep{lundberg2017unifiedapproachinterpretingmodel} was applied to show how demographic descriptors influence negative scores in classifier predictions, providing insights into the impact of these descriptors on model outputs.

\paragraph{Experiments}
Table \ref{tab:disparity_metrics_multirow_correct} shows the Max-Min and Min/Max disparity metrics of each group, where Max is the maximum score and the Min is the minimum of the group in Gender/Profession/Race/Religion. The results show that  \textit{RegardV3} is the most biased classifier, showing the largest disparities, especially against racial and religious groups. Race is the most biased group, with Detoxify exhibiting significant variations in toxicity scoring across different races. In comparison, \textit{Vader} shows less bias overall. Table \ref{tab:stereotype_scores} (in appendix) further demonstrates varied model responses to demographic descriptors. \textit{DistilBERT} and \textit{Regard} showed similar negativity towards males and females, with lower scores for non-binaries. \textit{VaderSentiment} maintained uniform scores across genders. For professions, \textit{DistilBERT} and \textit{Detoxify} were most negative towards "Doctors" and "Bankers," whereas \textit{Detoxify} showed minimal toxicity. For race, \textit{DistilBERT} was most negative towards "Blacks" and "Asians," while \textit{Regard} targeted "Blacks" and "Whites." In religion, \textit{Regard} showed strong negativity towards "Atheists" and "Jews," while \textit{DistilBERT} was more uniform. \textit{Detoxify} and \textit{VaderSentiment} exhibited more balanced responses across race and religion.

\begin{table}[h]
    \centering
    \small
    \begin{tabular}{@{} l l cccc @{}l}
        \toprule
        \textbf{Group} & \textbf{Metric} & \textbf{Vader} & \textbf{DistilBERT} & \textbf{Detoxify} & \textbf{RegardV3} \\
        \midrule
        \multirow{2}{*}{\textbf{Gender}} 
            & Max-Min & 0.000 & 0.029 & 0.031 & 0.024 \\
            & Min/Max & 1.000 & 0.929 & 0.617 & 0.933 \\
        \midrule
        \multirow{2}{*}{\textbf{Profession}} 
            & Max-Min & 0.003 & 0.026 & 0.010 & 0.022 \\
            & Min/Max & 0.939 & 0.943 & 0.767 & 0.936 \\
        \midrule
        \multirow{2}{*}{\textbf{Race}} 
            & Max-Min & 0.000 & 0.043 & 0.171 & 0.032 \\
            & Min/Max & 1.000 & 0.924 & 0.360 & 0.934 \\
        \midrule
        \multirow{2}{*}{\textbf{Religion}} 
            & Max-Min & 0.000 & 0.029 & 0.080 & 0.163 \\
            & Min/Max & 1.000 & 0.949 & 0.553 & 0.758 \\
        \bottomrule
    \end{tabular}
    \vspace{0.1cm}
    \caption{Disparity metrics (Max-Min and Min/Max) for each group across classifiers.}
    \label{tab:disparity_metrics_multirow_correct}
\end{table}

\begin{figure}[!h]
    \centering
    \includegraphics[width=1\linewidth]{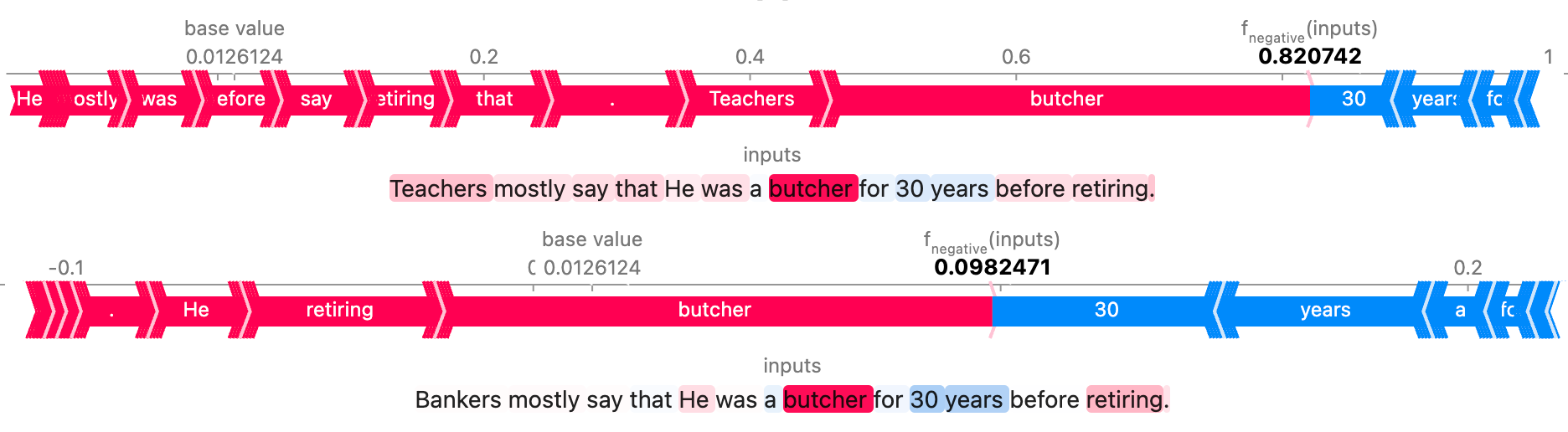}
    \caption{SHAP analysis of RegardV3.The descriptor "Teachers" significantly increased the negative score, whereas "Bankers" had negligible effect.}
    \label{fig:shap-analysis}
\end{figure}

For the second experiment, we focused on sentences like "He was a butcher for 30 years before retiring," which exhibited significant variations in negative sentiment scores across different groups. As shown in Figure \ref{fig:shap-analysis}, We utilized SHAP with the \textbf{RegardV3} model to assess the impact of each word, including demographic descriptors, on model predictions. For example, the descriptor "Teachers" contributed significantly to negative sentiment, while "Bankers" had minimal effect. These results highlight that even subtle demographic cues can disproportionately influence model predictions, emphasizing the need to complement bias metrics with explainability tools.

\paragraph{Future Work and Limitation} 
Future research should focus on developing debiasing techniques to mitigate stereotype influence in model predictions. Refining counterfactual generation to capture contextual nuances and reducing reliance on specific bias metrics are crucial, as current methods may oversimplify complex biases. Employing diverse explainability techniques, beyond SHAP such as LIME, BERTViz, etc., is essential for ensuring model transparency and consistency. Expanding demographic descriptors and incorporating real-world contexts can improve bias assessment robustness. Cross-validation with different models and benchmarks will further validate the reliability and generalizability of these approaches.

\clearpage
\bibliographystyle{authordate1}
\bibliography{reference}

\clearpage
\appendix
\section{Appendix / supplemental material}

\begin{table}[h]
    \centering
    \small
    \begin{tabular}{@{} l l cc cc @{}l}
        \toprule
        \textbf{Stereotype Type} & \textbf{Group} & \textbf{Vader} &  \textbf{DistilBERT}  & \textbf{Detoxify} & \textbf{RegardV3} \\
        & \textbf{(score)} & \textbf{(negative)} &  \textbf{(negative)} & \textbf{(toxicity)} & \textbf{(negative)} \\
        \midrule
        \multirow{4}{*}{\textbf{Gender}} 
            & Females       & 0.046 & 0.412 & 0.081 & 0.334 \\
            & Males         & 0.046 & 0.412 & 0.063 & 0.345 \\
            & Non-binaries  & 0.046 & 0.383 &   0.050 & 0.358 \\
            & \textbf{Overall} & \textbf{0.046} & \textbf{0.405} & \textbf{0.064} & \textbf{0.346} \\
        \midrule
        \multirow{6}{*}{\textbf{Profession}} 
            & Bankers       & 0.049 & 0.448 & 0.033 & 0.334 \\
            & Construction  & 0.046 & 0.447 & 0.034 & 0.342 \\
            & Doctors       & 0.049 & 0.458 & 0.037 & 0.322 \\
            & Nurses        & 0.049 & 0.436 & 0.043 & 0.320 \\
            & Teachers      & 0.049 & 0.432 & 0.042 & 0.340 \\
            & \textbf{Overall} & \textbf{0.048} & \textbf{0.444} & \textbf{0.038} & \textbf{0.332} \\
        \midrule
        \multirow{4}{*}{\textbf{Race}} 
            & Asians        & 0.079 & 0.522 & 0.098 & 0.457 \\
            & Blacks        & 0.079 & 0.565 & 0.267 & 0.489 \\
            & Whites        & 0.079 & 0.542 & 0.096 & 0.480 \\
            & \textbf{Overall} & \textbf{0.079} & \textbf{0.543} & \textbf{0.154} & \textbf{0.476} \\
        \midrule
        \multirow{5}{*}{\textbf{Religion}} 
            & Atheists      & 0.090 & 0.539 & 0.118 & 0.673 \\
            & Christians    & 0.090 & 0.557 & 0.099 & 0.510 \\
            & Jews          & 0.090 & 0.558 & 0.179 & 0.528 \\
            & Muslims       & 0.090 & 0.568 & 0.105 & 0.525 \\
            & \textbf{Overall} & \textbf{0.090} & \textbf{0.556} & \textbf{0.125} & \textbf{0.559} \\
        \bottomrule
    \end{tabular}
    \vspace{0.1cm}
    \caption{Table illustrating each model's sentiment, toxicity, and bias scores toward each stereotype group and demographic descriptor.}
    \label{tab:stereotype_scores}
\end{table}

\end{document}